\documentclass[letterpaper, 10 pt, conference]{IEEEtran}
\IEEEoverridecommandlockouts
\usepackage{graphics} % for pdf, bitmapped graphics files
\usepackage{graphicx}
\usepackage{epsfig} % for postscript graphics files
\usepackage{mathptmx} % assumes new font selection scheme installed
\usepackage{times} % assumes new font selection scheme installed
\usepackage{amsmath} % assumes amsmath package installed
\usepackage{amssymb}  % assumes amsmath package installed
\usepackage{listings}
\usepackage{subcaption}
\usepackage[hidelinks]{hyperref}
\usepackage{booktabs}       % For professional looking tables
\usepackage{colortbl}       % Define colored rows
\usepackage{array}          % For table wrapping and bold row
\usepackage{makecell}
\usepackage{mathtools}
\usepackage{yhmath}
\usepackage{amsfonts}
\usepackage{mathrsfs}
\usepackage{pifont}
\usepackage{mathpartir}
\usepackage[usenames,dvipsnames]{xcolor}
\usepackage{xspace}
\usepackage{ifdraft}
\usepackage{comment}
\usepackage{csquotes}
\usepackage{ragged2e}
\usepackage{textcase}
\usepackage{enumitem}
\usepackage{cleveref}
\usepackage{eccvabbrv}
\usepackage{placeins}
\newcommand{\vidtoreal}{\textsc{Vid2Real~HRI}\xspace}
\newcommand{\algoname}{\textsc{Vid2Real~HRI}}
\newcommand{\figspace}{\vspace{-10pt}}
\newcommand{\fwsection}[1]{\vspace{0.5em}\noindent\textbf{#1}}

\title{\large \bf
\algoname: Align video-based HRI study designs with real-world settings
}

\author{Elliott Hauser$^{1}$, Yao-Cheng Chan$^{1}$, Sadanand Modak$^{2}$, Joydeep Biswas$^{2}$, Justin Hart$^{2}$% <-this % stops a space
\thanks{*This research is part of an ongoing collaboration supported by NSF Award 2219236,  Living and Working with Robots, a core research project of Good Systems, a UT Grand Challenge, and an unrestricted gift from The MITRE Corporation. Dr. Hart is additionally supported by Cisco Research and Army Futures Command.}% <-this % stops a space
\thanks{$^{1}$School of Information, The University of Texas at Austin, {\tt\small \{eah13, ycchan\}@utexas.edu}}%
\thanks{$^{2}$Department of Computer Science, The University of Texas at Austin, {\tt\small \{sadanandm, joydeepb, justinhart\}@utexas.edu}}%
}

\begin{document}

\maketitle

\begin{abstract}
HRI research using autonomous robots in real-world settings can produce results with the highest ecological validity of any study modality, but many difficulties limit such studies' feasibility and effectiveness. We propose \algoname, a research framework to maximize real-world insights offered by video-based studies. The \algoname\ framework was used to design an online study using first-person videos of robots as real-world encounter surrogates. The online study ($n=385$) distinguished the within-subjects effects of four robot behavioral conditions on perceived social intelligence and human willingness to help the robot enter an exterior door.
A real-world, between-subjects replication ($n=26$) using two conditions confirmed the validity of the online study's findings and the sufficiency of the participant recruitment target ($22$) based on a power analysis of online study results. The \algoname\ framework offers HRI researchers a principled way to take advantage of the efficiency of video-based study modalities while generating directly transferable knowledge of real-world HRI. Code and data from the study are provided at \href{https://vid2real.github.io/vid2realHRI}{vid2real.github.io/vid2realHRI}.
\end{abstract}

\section{Introduction}
Human-Robot Interaction (HRI) research is intimately shaped by the nature of robots as technical artifacts. As robots are not part of daily life for most people around the world, studies of robots in real-world settings are thus limited to existing uses of robots or researcher-constructed scenarios, which are oriented towards validating new designs or robot behaviors \cite{Hauser2023-gf}, and often involve real-world designs where humans are state-of-the-art research robots. While these may gain direct applicability to expanding the frontiers of robotics, they often lose ecological validity \cite{Dole2019-yo}. The difficulties this dynamic presents call for a new research framework that is applicable to novel forms of autonomous robot HRI phenomena, such as fully autonomous robotics and incidental human-robot encounters (HRE) research in pedestrian settings.

HRI of autonomous robots in real-world settings presents the methodological challenge of deriving precise knowledge of a highly complex subject in a high-dimensional space of potential variables. Incidental encounters with robots, the scenario studied here, are fleeting yet increasingly common events with potentially profound social impact. Without directly addressing these challenges, the field of HRI risks delaying or forgoing the technical \cite{Biswas2021-gw} and theoretical \cite{Abrams2021-vw} opportunities of better understanding and reacting to them.

The present study proposes a new research framework, \algoname, to guide researchers studying HRI with autonomous robots in real-world settings seeking to use video-based studies. \algoname, described in \Cref{sec:framework}, allows researchers to better align video-based study designs with the real-world conditions they seek to understand.

We illustrate the application of the \algoname\ framework to an HRE research study on the effects of socially compliant autonomous robot behaviors on pedestrians. The results of the application study demonstrated that \algoname\  improved the video study's alignment with real-world settings, produced HRI knowledge with real-world applicability, and provides a compelling path towards specific follow-up studies. In addition to the statistical best practices it encourages \cite{Kang2021-cn}, the \algoname\ framework provides principled epistemological grounding for HRI research programs seeking to effectively and efficiently generate and build upon knowledge of complex sites, phenomena, and robotic systems.

\section{Related Work} \label{sec:relatedwork}

% Entering the methods convo
Amidst longstanding calls for standard metrics and heightened scientific rigor in the field of HRI \cite{Bethel2010-tk,Woods2006-dn}, an increasing number of researchers are conducting methodologically oriented studies, seeking to better understand and improve the scientific knowledge generated from HRI research practices. These have included studies of reproducibility \cite{Gunes2022-wt}, the relative validity of lab-based and online studies \cite{Babel2021-xf}, and the effects of different stimulus media, such as static images, moving images, and videos upon research results \cite{Randall2023-ra}.

% Research Frameworks proposed to address Characteristic Methods Difficulties
A range of frameworks have been advanced to guide research in ever-more situated and specialized contexts \cite{Rosenthal-von_der_Putten2021-pb, Damholdt2020-fr}. Kunold \cite{Kunold2022-vo}, for instance, notes a lack of principled ways for turning research insights into design principles as the rationale for a framework for studying communication with social robots. Frameworks have been advanced to guide research on complex topics like group dynamics \cite{Abrams2020-cd} or the formation of expectations of social robots \cite{Berzuk2023-jk}.

HRI with mobile autonomous robots in real-world settings faces a range of methodological difficulties arising from the high dimensionality of potential factors. Factors found to have influenced real-world HRI with autonomous robots range from the robot characteristics \cite{Liang2023-xc} to robot physical behavior \cite{Okafuji2022-wv} to human-robot communication \cite{Chen2021-ww}. While real-world HRI knowledge continues to advance, isolation and control of variables and context remains a substantial challenge, posing risks to the applicability of HRI research.

A potential solution to this problem is to use more highly controlled experimental setups in the laboratory or online \cite{Holman2021-pl}. The established validity of online studies for specific categories of HRI research questions \cite{Babel2021-xf,Randall2023-ra} is mitigated by its lack of fidelity of experience: seeing a video of an autonomous robot will rarely involve the full range of experiences present in real-world encounters. Laboratory studies, on the other hand, gain the ability to control variables and present in-person robotic stimuli to participants, but must trade the ecological validity of study setting inherent to research. Online studies have high statistical power and can be relatively well-controlled between participants, but have low ecological validity for studies of autonomous mobile robots specifically. Lab-based studies have the complex interactional possibilities of in-person studies, but trade statistical power to do so. Real-world studies' extremely high ecological validity is mitigated substantially. This situation, summarized in \Cref{tab:tradeoffs}, forces researchers to make difficult choices about the strengths and weaknesses of different study modalities. 

\begin{table*}
    \centering \setlist{left=0pt, topsep=0pt}
    \caption{Summary of tradeoffs between video-based study and real-world study.}
    \label{tab:tradeoffs}
    \begin{tabular}{ccc}
        \toprule
        % \cline{2-3}
        & \makecell*{\textbf{Video-based Online Study}} & \makecell*{\textbf{Real-world Study}} \\
        \midrule
        % \cline{2-3}
        \textsc{\textbf{Pros}} & 
         \makecell*[{{m{.35\linewidth}}}]{\begin{itemize}
             \item Reproducibility of stimulus
             \item Fast recruitment
             \item Ease of isolating factors
         \end{itemize}
         \textbf{Higher Statistical Power}} & 
         \makecell*[{{m{.35\linewidth}}}]{\begin{itemize}
             \item Multimodal data (e.g. sensors, observation)
             \item Longitudinal potential
             \item Site-specific context
         \end{itemize}\textbf{Higher Ecological validity}}\\
         % \cline{2-3}
         \hline
         \textsc{\textbf{Cons}} & 
         \makecell*[{{m{.35\linewidth}}}]{\begin{itemize}
            \item Skills for creating video encounters required
            \item Potential disconnection from specific contexts
            \item Homogenous data (e.g. no sensors, observation)
         \end{itemize}
         \textbf{Lower Ecological Validity}} & 
         \makecell*[{{m{.35\linewidth}}}]{\begin{itemize}
             \item Inherently Between subjects
             \item Labor-intensive
             \item Difficult to isolate factors
         \end{itemize}
         \textbf{Lower Statistical Power}}\\
         % \cline{2-3}
         \hline
         \textsc{\bfseries \makecell*{Example\\Studies}} 
         &
         \makecell*[{{m{.35\linewidth}}}]{
         Babel \etal \cite{Babel2021-xf},
         Randall and Sabanovic \cite{Randall2023-ra}} &
         \makecell[{{m{.35\linewidth}}}]{
            Liang \etal \cite{Liang2023-xc},
            Chen \etal \cite{Chen2021-ww}} \\
        % \cline{2-3}
        \bottomrule
    \end{tabular}
\end{table*}

Attempts to move outside the online/real-world dichotomy have yielded a range of partial solutions, including realistic laboratory settings, ad hoc incorporation of various modalities in a single study, or estimates of real-world validity based on published studies of very different design. While these approaches may increase real-world applicability, they can be difficult to evaluate and compare. HRI-specific study design frameworks responsive to the general move towards standardization \cite{Schrum2023-jl} in the field are needed to improve the efficiency and validity of real-world HRI studies of autonomous robots.

\section{\vidtoreal Framework} \label{sec:framework}
The \algoname\ framework offers HRI researchers a principled way to use the complementary strengths of video-based and real-world research modalities in studies such as Human-Robot Interaction or incidental Human-Robot Encounters with autonomous robots. It characterizes the methodological particularities of each modality for HRI studies of autonomous robots in real-world settings; and defines the epistemological relationships between them. This section describes the framework's rationale, design, and potential and recommended uses.

\subsection{Statistical and Epistemological Rationale}
% Statistics
Statistical best practice for performing a study requires a power analysis to inform the estimated sample size required to achieve the desired statistical power \cite{Bethel2010-tk,Kahng2019-of}. Past HRI research has utilized this practice, which becomes much easier if similar instruments are used for most studies. Wasted effort due to insignificant results is potentially avoidable through statistical techniques: accurate effect size estimates and power calculations can help inform researchers whether their design is likely to observe the intended effect. Statistical discipline, as urged by leading HRI methods reference sources \cite{Bethel2020-pi} dramatically increases the efficiency and quality of research activity.

% epistemology
However, statistical best practice isn't enough. Without a principled framework for when and how to conduct a study online or in the field, researchers risk a non-optimial blend of these modalities' tradeoffs, shown in \cref{tab:tradeoffs}. In contrast, the \algoname\ framework is designed to enhance real-world study outcomes by aligning the hypothesis and design of a video-based study with untested assumptions of a target real-world study. This alignment, described below and exemplified in \cref{sec:application}, is designed to achieve \textit{commensurable} results, wherein results from each study are informative and predictive of the other. \algoname\ thus takes an epistemological approach to the design of research studies, focusing on producing knowledge of unknowns and testing assumptions. In addition to facilitating statistical rigor, the framework enables a research team to efficiently and iteratively improve their knowledge of their subject and of their experimental intervention over a series of studies, within or across research publications. The \algoname\ framework's focus on alignment helps to identify and realize the synergistic value of commensurable results in single studies and within a broader research program. Video-based studies are used to produce highly specific knowledge that help explain and isolate phenomena of interest in real-world HRI studies. Unexpected results or events in real-world studies generate hypotheses for refinement in future online studies. The \vidtoreal framework thus helps researchers use their current state of knowledge  to determine which modality is most informative at a given stage of inquiry. 

\subsection{Framework Description}
\begin{figure}[htp]
    \figspace
    \centering
    \includegraphics[width=\columnwidth]{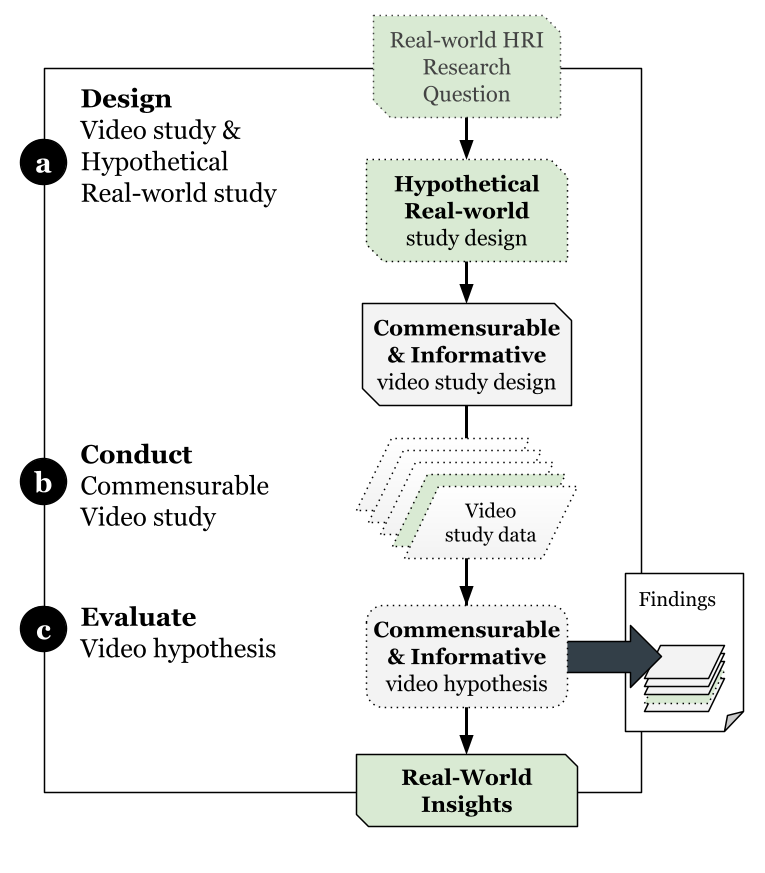}
    \caption{A schematic of the \algoname\ process}
    \label{fig:processdetail}
    \figspace
\end{figure}

\Cref{fig:processdetail} provides an overview of the \algoname\ framework. We specify that video-based study designs should be aligned with a specific real-world study's hypotheses, design, and scenario, regardless of whether such a real-world study is actually performed. Deliberate alignment with a real-world study design makes video-based study results commensurable with and informative about the real-world settings of interest. 

We define \emph{commensurable} as a relative property that obtains between video and real-world studies that produce directly comparable and mutually predictive results. The framework aims to acheive commensurability between video-based and real-world study designs by aligning variable(s), scenario, and settings. This alignment ensures that similar results are obtainable, in principle, in real-world settings. The degree of commensurability realized between two studies aligned in this way is primarily a function of the fidelity with which the video-based encounter surrogate evokes the real-world encounter of interest for study participants.

We define \emph{informative} as a property of the hypothesis tested in the video-based study, whereby the results have significant implications for real-world study design. The sense of information invoked here is from an information-theoretic sense: novel information is gained. Informativeness is, roughly, the expected surprise weighted by expected frequency. Informative study designs should thus test both assumptions, which are rarely expected to be violated but produce large amounts of novel information when they are, and empirical measurements, which produce small amounts of novel information reliably.

The \emph{hypothetical real-world study design} is defined as a version of the video-based study able to be conducted in person using the same variable(s), scenario, and setting. This study design is not carried out in all cases, would be carried out after the online results are obtained, and likely would be revised if carried out. The hypothetical real-world study design's primary function in the framework is as an alignment target for the video-based design, enabling the accomplishment of commensurability and informativeness. Follow-up real-world studies, if any, should have designs informed by the \algoname\ study results (otherwise, the real-world study's value would be primarily methodological, as they are in the present research). 

The following paragraphs reference and explain the \vidtoreal phase labels shown in \cref{fig:processdetail}.

\fwsection{a. Design.} Beginning with an HRI research question, researchers identify a \emph{hypothesis} and operationalize it in a \emph{hypothetical real-world study design} (\Cref{fig:processdetail}a). Researchers then develop a \emph{commensurable} video study hypothesis and study design, iteratively revising the designs as needed. The video stimulus is conceived as an \textit{encounter surrogate}, and can be judged by its capability to replicated participant experience intended in the hypothetical real-world study. The characteristics of videos may vary significantly by research question, so the \algoname\ framework does not specify standard methods. An example of creating videos for human-robot encounters research is provided in \Cref{sec:application}.

\fwsection{b. Conduct and c. Evaluate.} Once a commensurable video hypothesis and study design are ready and tested, the study can be conducted (\Cref{fig:processdetail}b). Note that standard best practices, such as obtaining institutional review board (IRB) approval and piloting the design should be followed, and are not shown in \Cref{fig:processdetail}. Due to the speed and relative cost-efficiency of participant recruitment, a large amount of data can be collected in a short amount of time. With this data collected, researchers can evaluate the commensurable video study hypothesis (\Cref{fig:processdetail}c), producing findings (these may be publishable or contribute to future publications, depending on their significance).

The findings of \algoname\ study can of course be published without real-world replication if they are significant and novel. The insights produced by the alignment of study design with a specific real-world setting aid in interpretation of findings, the selection of future research questions, and design of real-world of video-based studies of the same or related real-world settings and scenarios. When warranted, the hypothetical real-world study design may be modified in light of the online video study results, supplementing the original results or constituting an independent research contribution. Conducting the hypothetical real-world study design with little modification, as done in the present paper, can characterize the methodological validity of the framework in specific applications where the applicability of the framework's validity (demonstrated below) is uncertain.

\section{\vidtoreal Online Study} \label{sec:application}
The \vidtoreal framework was applied to a specific HRE research study on the effects of socially compliant robot behavior conditions, operationalized as mixtures of verbal cues and body language, on pedestrian perceptions of the robot's social intelligence and willingness to open a door for the robot.
The application study's illustrative value and relevance to the strengths of the \vidtoreal framework is the focus of its presentation in this paper; full analysis of results and their implications for HRE research is presented elsewhere.

\subsection{Application Online Study Design}
The study scenario is an incidental encounter with a quadruped robot requiring human assistance to enter a building. Specifically, the robot is assisted by a human pushing a button to automatically open the doors. Since the nearby human has no obligation to assist the robot, we hypothesized that increased perceived social intelligence would increase human compliance with the robot's need for assistance.

Four robot behavioral conditions operationalized differentially socially compliant behaviors using verbal cues and body language:
\begin{itemize}
    \item Baseline:
    The walking robot stops at the entrance and waits for the nearby human to push the button, without offering any indications that it needs assistance.
    \item Verbal:
    The walking robot stops at the entrance and says, \emph{\enquote{Excuse me, can you open the door for me?}} to the nearby human, then waits for the human to push the button.
    \item Body Language:
    The walking robot stops at the entrance and turns its head facing the nearby human, then turns its head back to look at the door, then waits for the human to push the button.
    \item Body Language + Verbal:
     The walking robot stops at the entrance and turns its head facing the nearby human and says, \emph{\enquote{Excuse me, can you open the door for me?}}, then turns its head back to look at the door and waits for the human to push the button.
\end{itemize}

\begin{figure}[htp]
    \figspace
    \centering
    \includegraphics[width=\columnwidth]{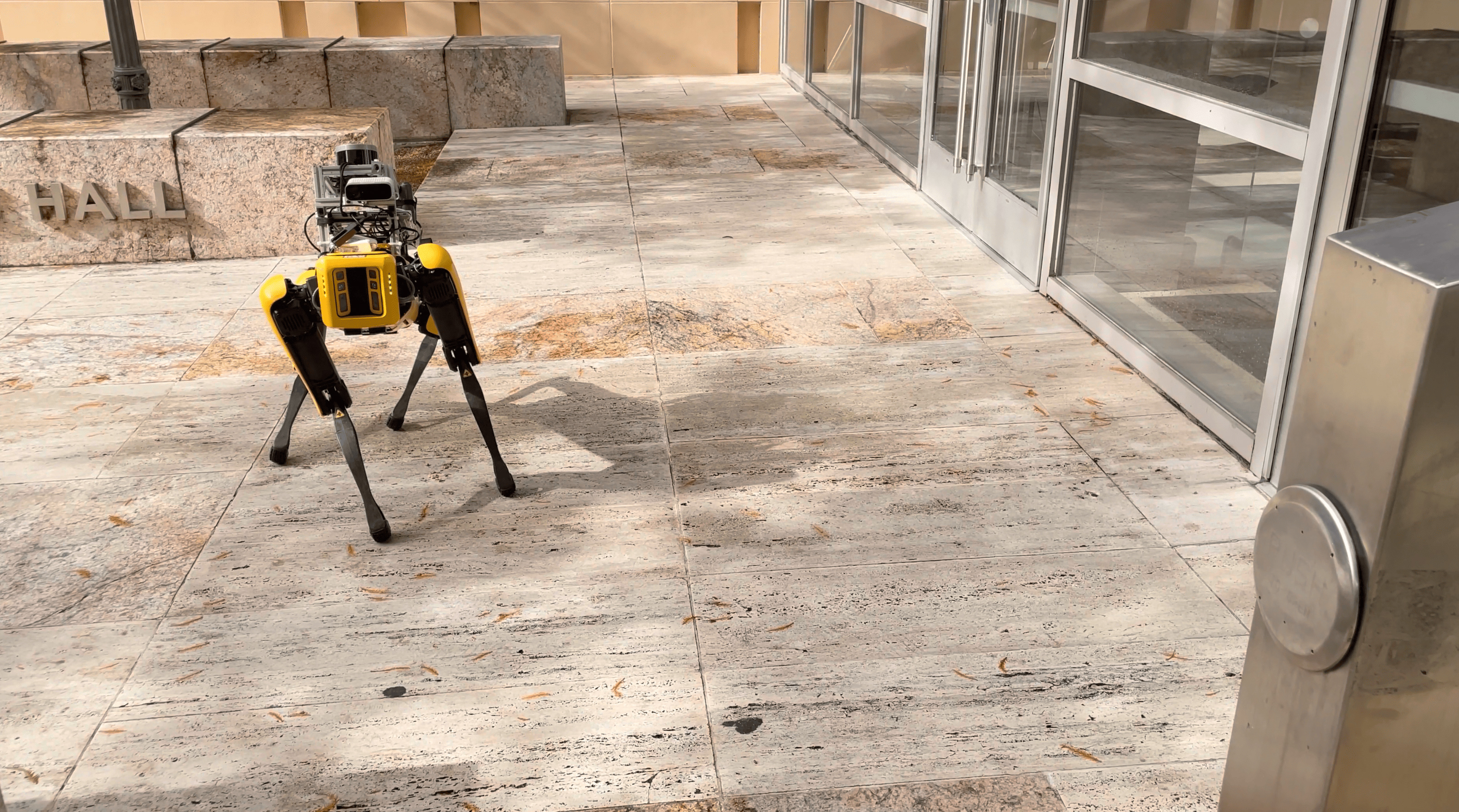}\\
    \vspace{0.2cm}
    \includegraphics[width=\columnwidth]{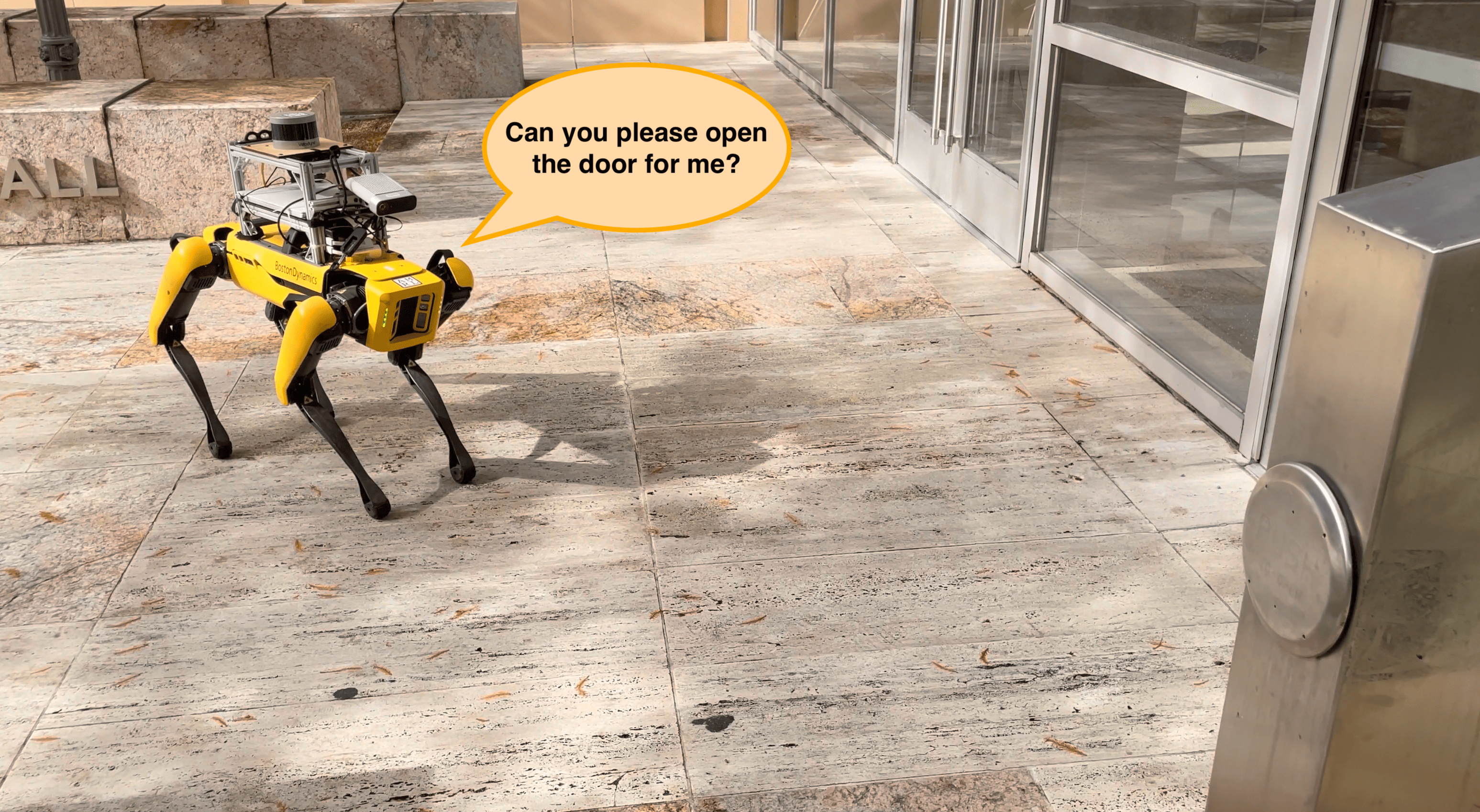}
    \caption{Sample frames from the videos of the Body Language \textbf{(top)} and Verbal \textbf{(Bottom)} conditions.}
    \label{fig:videofigures}
    \figspace
\end{figure}

\subsubsection{Video and Real-World Alignment}
To ensure the alignment of the video settings and the real-world scenarios, the videos are shot from a first-person point-of-view. This decision is mainly to support the need for the participants to imagine that they are the person encountering the robot. The robot motion is also reproduced across the conditions in order to maintain consistency. Additionally, the location where the videos are shot is the same as the place where the future in vivo HRE research will be conducted. Therefore, the ecological validity of this online study is reserved, and the results are commensurable. The videos and a detailed description of the study design are publicly available \cite{Chan2024-ad}.

\subsubsection{Questionnaire}
The Perceived Social Intelligence (PSI) scale \cite{Barchard2020-ug} was adopted as the measurement of the effects of robot body language and verbal cues. According to the authors' suggestions, research can adopt a subset of PSI that suits the research context and goals. We selected 8 of PSI's 13 Social Information Processing scales: (1) Social Competence, (2) Identifies Humans, (3) Identifies Individuals, (4) Recognizes Human Behaviors, (5) Adapts to Human Behaviors, (6) Predicts Human Behaviors, (7) Recognizes Human Cognitions, and (8) Adapts to Human Cognitions. Other scales were not chosen because they were not applicable to a first-person point-of-view video-based study.

In addition to PSI, two exploratory questions were presented with each study condition. The first was a five-point Likert question, "The person in the video represents how you would have responded in the situation", which was included to assess the participants' perceptions of the fidelity of the human's actions in the video to their own. The second was a free-form text response question that asked the participants to answer how they would have behaved in the situation that they just watched. 

\subsubsection{Study procedure}
The participants were recruited on Prolific. They had to have no visual and hearing difficulties, a high level of English fluency, at least a 95\% approval rate on Prolific, and currently live in the United States. The online study used a within-subject design, and the order of the study conditions was randomized to avoid the learning effect. The participants had to first consent to participate, and then they would be presented with one study video and the questionnaire. This step was randomly repeated four times. The total study lasted around 10 minutes. The University of Texas at Austin IRB reviewed and approved this study.

\subsection{Application Study Results}
\subsubsection{Participants}
The dataset of the application online study is publicly available \cite{Chan2024-ad} for review and reuse. The Prolific platform recruited 420 participants. However, upon careful investigation of response quality, 35 participants were excluded due to incomplete answers or cheating behaviors, leaving 385 valid responses for result analyses. Among the 385 participants, 194 participants identified as female, and 191 as male, the age ranged from 19 to 75 (\textit{M} = 38.53, \textit{SD} = 12.86).

\subsubsection{Human Compliance}
The human compliance was measured using the 5-point Likert exploratory question reported above. The average score are as follows: Baseline \textit{M} = 3.371, \textit{SD} = 1.35, Verbal: \textit{M}= 4.283, \textit{SD} = 1.038, Body Language \textit{M} = 3.698, \textit{SD} = 1.281, Body Language + Verbal \textit{M} = 4.376, \textit{SD} = 0.992. Using the one-way repeated measures ANOVA test with Bonferroni correction (corrected significance level is 0.008333)), the results showed that there was a significant main effect (\textit{F}(3, 1552) = 129.88, \textit{p} \textless 0.001, \(\eta_p^2\) = 0.252). Pairwise comparisons were also used as ad-hoc analyses. The results showed that only the pair of Body Language + Verbal condition and Verbal condition did not have a significant difference (\textit{p} = 0.03497). Based on these findings, it was clear that adding body language and verbal commands were efficient ways of increasing human compliance in an HRE context. Meanwhile, the verbal command had a significantly stronger effect than the body language.

\subsubsection{PSI ratings}
Given the within-subject design, the one-way repeated measures ANOVA test with Bonferroni correction was used to evaluate the differences. The Body Language + Verbal condition was rated the highest (\textit{M} = 3.826, \textit{SD} = 0.661), followed by the Verbal condition (\textit{M} = 3.611, \textit{SD} = 0.657), followed by the Body Language condition (\textit{M} = 3.601, \textit{SD} = 0.595), and the Baseline condition was rated the lowest (\textit{M} = 3.151, \textit{SD} = 0.784). The ANOVA test showed a main effect on PSI overall rating  (\textit{F}(3, 1552) = 144.26, \textit{p} \textless 0.001, \(\eta_p^2\) = 0.272), indicating the robot's socially compliant behaviors had positive influences on its perceived social intelligence. Additionally, higher PSI seemed to indicate higher human compliance. Therefore, the higher socially intelligent a robot is, the more likely humans would comply and offer help when the robot asks for one. Please note that full analyses, such as pairwise comparisons and ANOVA tests on the subscales, are omitted from the scope of this study.

\begin{table*}[ht]
    \centering
    \caption{Power analyses and the estimated \emph{n}}
    \label{tab:effectsize}
    \begin{tabular}{@{}llccc@{}}
        \toprule
        \bfseries Intervention & \bfseries Control & \bfseries \( n \) for \( \beta = .95 \) & \bfseries \( \eta_p^2 \) & \bfseries Needed increase of \( n \) \\ 
        \midrule
        Body language + Verbal & Baseline & 22 & 0.413 & 0.00\% \\
        Verbal & Baseline & 32 & 0.3043 & 45.45\% \\
        Body language & Baseline & 36 & 0.2782 & 63.64\% \\
        Body language + Verbal & Body language & 70 & 0.164 & 218.18\% \\
        Body language & Verbal & 6902 & 0.0019 & 31,272.73\% \\ 
        \bottomrule
    \end{tabular}
\end{table*}

\subsubsection{Power analyses}
An initial contribution of the \vidtoreal framework is a reliable estimation of the number of participants needed to observe a significant effect.  Real-world studies of incidental human-robot encounters almost always require between-subject designs due to the nature of the scenario. Between-subject designs typically require a larger sample size than within-subject designs, threatening researchers' ability to achieve their research goals. Drawing from the average of the 8 PSI scale ratings, power analyses of the Baseline condition compared to other conditions, and the Body Language condition compared to the Body Language + Verbal condition were conducted. The results are summarized in \Cref{tab:effectsize}. The results showed that 22 people would be required to reach a power of 0.95 in a between-subject study of Baseline and Body Language + Verbal condition. We also conducted an analysis of the Body Language condition compared to the Verbal condition, and the result was that 6902 people (${F}(1,384) = 0.723$, ${p} < 0.395$, \(\eta_p^2 = 0.00188\)) would be required. Please note that a between-subject design is chosen because it can better reflect the natural settings of real-world scenarios than a within-subject design.

\section{Real-World study}
\subsection{Study Design}
Following the \vidtoreal framework and the knowledge obtained from the online video study, the Baseline and Body Language + Verbal conditions with a between-subject design were chosen for the real-world study due to its achievable sample size. The real-world study took place at the exact same location where the videos were shot. The participants were instructed to start at the same location as in the video and walk in the same direction. After encountering the robot, the participants had the freedom to do whatever they felt natural as pedestrians who incidentally encountered the robot. After this interaction, the participants had to answer a questionnaire, which contained the same items, except for the two exploratory questions, which were excluded as they were not applicable. In replacement of the exploratory questions, a camera was set up to record the interaction session, specifically focused on how the participants reacted to the robot (\ie, whether or not they complied).

The robot behaviors were also the same as in the video studies. However, given the challenges of achieving full robot autonomy in such a real-world scenario, and more importantly to ensure that the robot behaviour is as close as possible to the video, the robot was teleoperated using a Wizard-of-Oz method. One author of the present study was designated as the study manager, who faced and gave instructions to the participants and pretended to start the study session using a laptop that was visible to them, while another author of this paper teleoperated the robot staying at a place where they could perfectly view the study process while being hidden from the participants. The process and setup were validated with pilot tests. The real-world study was approved by The University of Texas at Austin IRB. Upon successful completion of the study, all participants were entered into a draw to stand a chance to win a \$100 gift card, with one of them finally receiving the gift card.

\subsection{Study Result}
A total of 26 participants were recruited, 13 experienced the Baseline condition and 13 had the Body Language + Verbal condition, 13 identified as males, 13 as females. The age ranged from 18 to 28 (\textit{M} = 21.04, \textit{SD} = 3.01). Given the sample size and the between-subject design, the Mann-Whitney U test was used to compare the PSI ratings between the conditions (Baseline: \textit{M} = 3.433, \textit{SD} = 0.680, Body Language + Verbal: \textit{M}= 3.952, \textit{SD} = 0.630). As expected, the results showed a significant difference (\textit{z} = 1.9801, \textit{p} = 0.047, \textit{r} = 0.39). In terms of human compliance, 11 out of 13 participants opened the door in the Body Language + Verbal condition, whereas only one participant did so in the Baseline condition. Using Fisher's exact test, the Body Language + Verbal condition had a significantly higher proportion of participants who opened the door (\textit{p} \textless 0.001).

\section{Discussion}
\subsection{{\vidtoreal} Benefits in the Application Study}
The benefits of the \vidtoreal's Online Video modality on the application HRE study are demonstrated through the data results and the accumulated knowledge gained in the study material preparation process.

\subsubsection{Aligned results}
The video study's statistical results serve as both independent findings and empirical foundations for future in vivo research of autonomous robot HRI. The \vidtoreal process of designing a video-based study commensurable with a target real-world study shown in \cref{fig:processdetail} enabled this advantage by keeping real-world application a central concern in video study design.  Aligning an online video study with a specific real-world study design is an effective way to disentangle the high-dimensionality nature of HRI research of autonomous robots in real-world settings. Future HRE research may also build upon the findings obtained from online video studies and expand its scope to examine other factors underlying social environments.

The results of the PSI scale comparisons and the power analyses are sources for condition selection, especially for the between-subject design nature of research in real-world settings. For instance, in a research context the same as the application study, we learn that a study able to use the Baseline and Body Language + Verbal conditions results for effect size estimation leads to a reasonable sample size ($n=22$). This could inform future real-world studies or enable researchers to confidently select a multi-condition $N x M$ design without losing statistical significance by dividing their online study population too finely.

Small effect sizes or non-significance between conditions revealed by our approach were also informative. A comparison between the effect of body language and verbal cues on pedestrian perceptions would be drastically informed by the small effect size, the ${p}$ value, and the extremely large sample size ($n=6,902$) implied by these results. This suggests that a new study design would be required to distinguish the effects of body language and verbal forms of socially compliant behavior upon human compliance. A \vidtoreal study could rapidly inform selection of behavioral conditions suitable for investigating such a question.

\subsection{Optimizing Modality Strengths}
\vidtoreal helped us navigate the strengths and weaknesses of video and real-world study modalities (in \cref{tab:tradeoffs}). This section surveys our approach and highlights insights for researchers seeking to adopt the \vidtoreal framework in their research.

Preparing the video encounter surrogate in situ helped us refine the study design and proactively control the eventual real-world encounters. Our first videos revealed that pedestrians might not realize that the robot is attempting to have their attention, the door might be opened from inside, pedestrians might walk in front of the robot or follow it without interacting with it (in our case, offering help). During the real-world study, several of these issues happened again, making maintaining identical behavior across all study sessions difficult. These led to small variations to the location and timing of the robot's behavior. Participants' pace also naturally varied as well, causing the robot to miss the opportune moment to perform its behaviors. Some participants didn't see the body language, instead only hearing the verbal command. These exigencies are components of real-world studies' ecological validity, but could obscure valid findings if they obscure the effects of the study conditions. Confirming the effect using online study before entering the field helped us anticipate and mitigate these potential confounds through study design, while the power analysis helped us scope the study with confidence. The seemingly `methodologically better' alternative, starting with a real-world study, would have risked selecting conditions requiring infeasible sample sizes, resorting to a between-subjects design (diminishing ecological validity), and/or recruiting too few participants.

We view our online study videos as \textit{encounter surrogates}, aiming to replicated the conditions of incidental encounters with high fidelity. We thus used a first-person perspective of the encounter, rather than a third-person view of someone encountering a robot. This removed the variable of the appearance of the human as a potential confound, and the possibility of the participants trying to interpret the human actor's behaviors and be biased. Additionally, the questionnaire instructed participants to imagine that they were the person encountering the robot. 

\subsection{Summary of Framework Benefit}
The \vidtoreal research framework offered a principled way to design a video-based study of encounters with a quadruped robot aligned to a specific real-world setting. The online study's alignment with a specific real-world study design enabled it to efficiently produce valid knowledge of real-world encounters in a specific site. The online study results produced power analysis-based estimates of required sample size that accurately represented the statistical power obtained in the real-world study. Finally, the researchers' experiential insight into the importance of the direct `gaze' of the robot for the effectiveness of body language cues as perceived by humans suggests limitations of the chosen platform that can be directly targeted in future research. Together, these improvements increased the validity of application study results to a specific real-world setting and clarified the ways the results can directly inform future work.

\subsection{Limitations and Future Work}
While our real-world results suggest that our encounter surrogate was sufficient to predict real-world encounter data, future research is needed to understand effects of video different techniques upon surrogates' fidelity. 
This study only used video-based encounter surrogates due to their simplicity and suitability for HRE research. Video-based surrogates will likely be less informative to studies involving extensive user interaction with robots. Interactive study modalities such as simulation and VR/AR experiences may be effective encounter surrogates, but were not evaluated in this study.

\section{Conclusion} \label{sec:conclusion}
\vidtoreal is a research framework that leverages the complementary benefits of video-based and real-world autonomous robot HRI study designs. The \vidtoreal framework was used to design an online, video-based study aligned with a real-world study design in a specific site. 
The online study videos, conceived of as \textit{encounter surrogates}, utilized a first-person video of autonomous behavior conditions planned for the real-world study. This validated key study assumptions about the body language and verbal conditions' effectiveness, provided valuable characterization of the chosen survey scales over potential alternatives, and enabled targeted refinement of the autonomy. The aligned online and real-world study designs produced commensurable and informative findings, as demonstrated by similarity of results and accuracy of sample size estimates. The framework's benefits compared favorably to potential alternatives, such as lab-based studies or longitudinal in-situ instrumentation of real-world sites. The \vidtoreal framework offers HRI researchers a principled way to use the complementary strengths of video-based and real-world research modalities to understand and improve real-world encounters with autonomous robots.

\bibliographystyle{IEEEtran}
\bibliography{eh-paperpile,yc-paperpile,sm-refs}

\appendix
\subsection{Pairwise Comparisons of Conditions by PSI Scores}
\Cref{tab:pairwiseAB,tab:pairwiseAC,tab:pairwiseIH,tab:pairwiseII,tab:pairwisePB,tab:pairwiseRB,tab:pairwiseRC,tab:pairwiseAll,tab:pairwiseSOC} show the pairwise comparison of different interventions based on PSI scores.
\begin{table}[tbhp]
    \centering
    \caption{Pairwise comparison - Social Competence}
    \label{tab:pairwiseSOC}
    \begin{tabular}{cccc}
        Pair&  Mean Difference&  F statistic& p-value\\\hline
        Base-Verb&  0.885&  226.977& \textless0.001\\
        Base-Body&  0.348&  18.212& \textless0.001\\
        Base-BL+V&  1.226&  401.571& \textless0.001\\
        Verb-Body&  0.537&  51.017& \textless0.001\\
        Verb-BL+V&  0.340&  45.713& \textless0.001\\
        Body-BL+V&  0.877&  118.539& \textless0.001\\
    \end{tabular}
\end{table}

\begin{table}[tbhp]
    \centering
    \caption{Pairwise comparison - Identifies Humans}
    \label{tab:pairwiseIH}
    \begin{tabular}{cccc}
         Pair&  Mean Difference&  F statistic& p-value\\\hline
         Base-Verb&  0.802&  190.848& \textless0.001\\
         Base-Body&  0.831&  177.925& \textless0.001\\
         Base-BL+V&  1.018&  258.512& \textless0.001\\
         Verb-Body&  0.028&  0.481& 0.488\\
         Verb-BL+V&  0.215&  34.166& \textless0.001\\
         Body-BL+V&  0.187&  28.961& \textless0.001\\
    \end{tabular}
\end{table}

\begin{table}[tbhp]
    \centering
    \caption{Pairwise comparison - Identifies Individuals}
    \label{tab:pairwiseII}
    \begin{tabular}{cccc}
         Pair&  Mean Difference&  F statistic& p-value\\\hline
         Base-Verb&  0.582&  113.824& \textless0.001\\
         Base-Body&  0.571&  93.503& \textless0.001\\
         Base-BL+V&  0.797&  163.145& \textless0.001\\
         Verb-Body&  0.010&  0.046& 0.829\\
         Verb-BL+V&  0.215&  24.443& \textless0.001\\
         Body-BL+V&  0.226&  27.822& \textless0.001\\
    \end{tabular}
\end{table}

\begin{table}[tbhp]
    \centering
    \caption{Pairwise comparison - Recognizes Human Behaviors}
    \label{tab:pairwiseRB}
    \begin{tabular}{cccc}
         Pair&  Mean Difference&  F statistic& p-value\\\hline
         Base-Verb&  0.316&  47.377& \textless0.001\\
         Base-Body&  0.351&  58.324& \textless0.001\\
         Base-BL+V&  0.454&  86.911& \textless0.001\\
         Verb-Body&  0.033&  0.718& 0.397\\
         Verb-BL+V&  0.137&  14.029& \textless0.001\\
         Body-BL+V&  0.104&  9.507& 0.002\\
    \end{tabular}
\end{table}

\begin{table}[tbhp]
    \centering     
    \caption{Godspeed~\cite{Bartneck2009-js} questions and response types}
    \label{tab:questionnairegod}
    \begin{tabular}{rll}
        \toprule
        \multicolumn{1}{r}{\bfseries Category} & \bfseries Question & \bfseries Response \\
        \midrule
        \multicolumn{1}{r}{Anthropomorphism} & Fake - Natural & 6 pt. bipolar \\ 
        & Machinelike - Humanlike & 6 pt. bipolar \\ 
        & Unconscious - Conscious & 6 pt. bipolar \\ 
        & Artificial - Likelike & 6 pt. bipolar \\ 
        & \makecell[tl]{Moving rigidly -\\Moving elegantly} & 6 pt. bipolar \\
        \bottomrule
    \end{tabular}
\end{table}

\begin{table}[tbhp]
    \centering
    \caption{Pairwise comparison - Adapts to Human Behaviors}
    \label{tab:pairwiseAB}
    \begin{tabular}{cccc}
         Pair&  Mean Difference&  F statistic& p-value\\\hline
         Base-Verb&  0.330&  35.193& \textless0.001\\
         Base-Body&  0.270&  25.814& \textless0.001\\
         Base-BL+V&  0.478&  75.702& \textless0.001\\
         Verb-Body&  0.059&  1.535& 0.216\\
         Verb-BL+V&  0.148&  9.244& 0.003\\
         Body-BL+V&  0.207&  22.368& \textless0.001\\
    \end{tabular}
\end{table}

\begin{table*}[tbhp]
    \centering
    \caption{PSI~\cite{Barchard2020-ug} questionnaire questions and response type}
    \label{tab:questionnairepsi}
    \begin{tabular}{rp{.5\linewidth}l}
        \toprule
        \multicolumn{1}{r}{\bfseries Scale} & \bfseries Question & \bfseries Response \\
        \midrule
        \multicolumn{1}{r}{Social Competence (SOC)} & This robot is socially competent & 5 pt. Likert \\ 
        \multicolumn{1}{r}{Identifies Humans (IH)} & This robot notices human presence & 5 pt. Likert \\
        \multicolumn{1}{r}{Identifies Individuals (II)} & This robot recognizes individual people. & 5 pt. Likert \\ 
        \multicolumn{1}{r}{Recognizes Human Behaviors (RB)} & This robot notices when people do things. & 5 pt. Likert \\ 
        \multicolumn{1}{r}{Adapts to Human Behaviors (AB)} & This robot adapts effectively to different things people do. & 5 pt. Likert \\ 
        \multicolumn{1}{r}{Predicts Human Behaviors (PB)} & This robot anticipates people’s behavior. & 5 pt. Likert \\ 
        \multicolumn{1}{r}{Recognizes Human Cognitions (RC)} & This robot can figure out what people think. & 5 pt. Likert \\ 
        \multicolumn{1}{r}{Adapts to Human Cognitions (AC)} & This robot adapts its behavior based upon what people around it know. & 5 pt. Likert \\ 
        % \multicolumn{1}{r}{} &  &  \\
        \arrayrulecolor{lightgray} % Set the color of the upcoming line to gray
        \midrule
        \arrayrulecolor{black} % Reset the color back to black for future lines
        \multicolumn{1}{r}{\bfseries Exploratory questions} &  &  \\
        \midrule
        \multicolumn{1}{r}{Compliance likelihood} & The person in the video represents how you would have responded in the situation. & 5 pt. Likert \\ 
        \multicolumn{1}{r}{} & How would you have behaved in this situation? & Free Text \\
        \bottomrule
    \end{tabular}
\end{table*}

\begin{table}[tbhp]
    \centering
    \caption{Pairwise comparison - Predicts Human Behaviors}
    \label{tab:pairwisePB}
    \begin{tabular}{cccc}
         Pair&  Mean Difference&  F statistic& p-value\\\hline
         Base-Verb&  0.143&  5.415& 0.021\\
         Base-Body&  0.379&  45.561& \textless0.001\\
         Base-BL+V&  0.358&  34.13& \textless0.001\\
         Verb-Body&  0.236&  18.793& \textless0.001\\
         Verb-BL+V&  0.216&  17.303& \textless0.001\\
         Body-BL+V&  0.021&  0.177& 0.6739\\
    \end{tabular}
\end{table}

\begin{table}[tbhp]
    \centering
    \caption{Pairwise comparison - Recognizes Human Cognitions}
    \label{tab:pairwiseRC}
    \begin{tabular}{cccc}
         Pair&  Mean Difference&  F statistic& p-value\\\hline
         Base-Verb&  0.259&  24.09& \textless0.001\\
         Base-Body&  0.400&  68.286& \textless0.001\\
         Base-BL+V&  0.413&  52.605& \textless0.001\\
         Verb-Body&  0.140&  7.086& 0.008\\
         Verb-BL+V&  0.153&  9.186& 0.003\\
         Body-BL+V&  0.013&  0.063& 0.802\\
    \end{tabular}
\end{table}

\begin{table}[tbhp]
    \centering    
    \caption{Pairwise comparison - Adapts to Human Cognitions}
    \label{tab:pairwiseAC}
    \begin{tabular}{cccc}
         Pair&  Mean Difference&  F statistic& p-value\\\hline
         Base-Verb&  0.368&  37.269& \textless0.001\\
         Base-Body&  0.340&  36.078& \textless0.001\\
         Base-BL+V&  0.522&  71.693& \textless0.001\\
         Verb-Body&  0.028&  0.274& 0.601\\
         Verb-BL+V&  0.153&  8.387& 0.004\\
         Body-BL+V&  0.181&  12.954& \textless0.001\\
    \end{tabular}
\end{table}

\begin{table}[tbhp]
    \centering    
    \caption{Pairwise comparison - All}
    \label{tab:pairwiseAll}
    \begin{tabular}{cccc}
         Pair&  Mean Difference&  F statistic& p-value\\\hline
         Base-Verb&  0.461&  167.924& \textless0.001\\
         Base-Body&  0.436&  147.988& \textless0.001\\
         Base-BL+V&  0.658&  271.385& \textless0.001\\
         Verb-Body&  0.024&  0.723& 0.395\\
         Verb-BL+V&  0.197&  50.745& \textless0.001\\
         Body-BL+V&  0.222&  75.328& \textless0.001\\
    \end{tabular}
\end{table}

\begin{figure}[tbhp]
    \figspace
    \centering
    \includegraphics[width=\columnwidth]{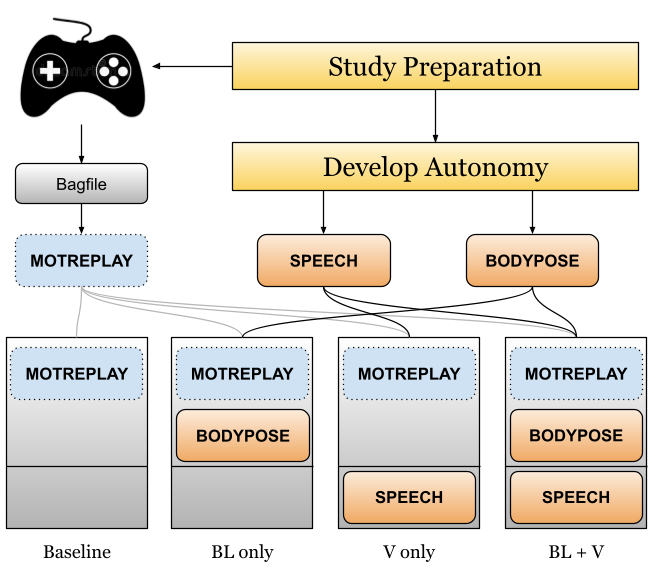}
    \caption{Overview of the methodology for recording videos}
    \label{fig:studyprep}
    \figspace
\end{figure}

\subsection{Questionnaire 1: PSI}
\Cref{tab:questionnairepsi} shows the questions chosen from the PSI questionnaire's various scales, as well as the two exploratory questions that we added.

\subsection{Questionnaire 2 - Godspeed}
\Cref{tab:questionnairegod} shows the questions chosen from the Godspeed questionnaire.

\subsection{Questionnaire 3 - AMPH}
\Cref{tab:questionnaireamph} shows the questions chosen from the AMPH questionnaire.

\subsection{Video Recording Methodology for Online Video Study}
For the purposes of this study, the Wizard-of-Oz design was used. More specifically, the robot operation was semi-autonomous, with the navigation being manually teleoperated by an operator. These modules supplemented manual operation:
\begin{itemize}
    \item \textsc{Speech}: This uses the python libraries (gTTs and pydub) to produce audio from text on the fly.
    \item \textsc{Bodypose}: Uses pointcloud-based object detection \cite{pcdet2020} to detect the observer and leverages the Clearpath Spot-ROS wrapper (built upon Boston Dynamics API) to modify the pose of the robot in a way that its head is ``looking'' at the observer.
    \item \textsc{Motreplay}: It uses a recorded bagfile and republishes the stream of motion commands required to generate identical navigation motion for the robot.
\end{itemize}

\begin{table*}[thbp]
    \centering     
    \caption{AMPH~\cite{amphquestionnaire} questionnaire questions and response type}
    \label{tab:questionnaireamph}
    \begin{tabular}{rp{.5\linewidth}l}
        \toprule
        \multicolumn{1}{r}{\bfseries AMPH} & \bfseries Question & \bfseries Response \\
        \midrule
        & Would you name an everyday object? & 4 pt. Likert-like \\ 
        & Do you ever blame your computer for being uncooperative? & 4 pt. Likert-like \\ 
        & Do you believe that sometimes your computer sabotages your actions on purpose? & 4 pt. Likert-like \\ 
        & Do you find it understandable when people treat their car as if it were human? I.e. by giving it a name and referring to it as \`{}reliable\`{} and \`{}helpful\`{}? & 4 pt. Likert-like \\ 
        & Do you tend to feel grateful toward a technological object (your car, your computer, your mobile) if it has rescued you from a dangerous or difficult situaiton? & 4 pt. Likert-like \\ 
        & Would you ever believe that a mountain (such as Mount Everest) would set off a series of avalanches on mountain climbers if they disturb the peace of the mountain? & 4 pt. Likert-like \\ 
        & Do you believe that a tree can feel pain? & 4 pt. Likert-like \\ 
        & Do you believe that an insect has a soul that you need to respect? & 4 pt. Likert-like \\
        \bottomrule
    \end{tabular}
\end{table*}

Figure \ref{fig:studyprep} shows the high-level overview of the methodology followed for recording videos. To make the different conditions perfectly comparable, the videos were shot in a way that preserves this consistency between the videos. To do so, robot motion was reproduced across different conditions. The main idea was to record a bagfile with the relevant rostopics while the robot performs the baseline behavior, and then use \textsc{Motreplay} to generate the same motion while we add other required behaviors depending on the condition we were recording the video for on the fly.
\FloatBarrier
However, due to differing robot behaviors among the conditions, the Baseline condition and Verbal condition used one bagfile, while the Body Language condition and the Body Language + Verbal condition used another bagfile as they required additional behaviors. 

\end{document}